# Large-Scale Traffic Data Imputation with Spatiotemporal Semantic Understanding

Kunpeng Zhang, Lan Wu, Liang Zheng, Na Xie, and Zhengbing He, *Senior Member, IEEE*

*Abstract*—Large-scale data missing is a challenging problem in Intelligent Transportation Systems (ITS). Many studies have been carried out to impute large-scale traffic data by considering their spatiotemporal correlations at a network level. In existing traffic data imputations, however, rich semantic information of a road network has been largely ignored when capturing network-wide spatiotemporal correlations. This study proposes a Graph Transformer for Traffic Data Imputation (GT-TDI) model to impute large-scale traffic data with spatiotemporal semantic understanding of a road network. Specifically, the proposed model introduces semantic descriptions consisting of network-wide spatial and temporal information of traffic data to help the GT-TDI model capture spatiotemporal correlations at a network level. The proposed model takes incomplete data, the social connectivity of sensors, and semantic descriptions as input to perform imputation tasks with the help of Graph Neural Networks (GNN) and Transformer. On the PeMS freeway dataset, extensive experiments are conducted to compare the proposed GT-TDI model with conventional methods, tensor factorization methods, and deep learning-based methods. The results show that the proposed GT-TDI outperforms existing methods in complex missing patterns and diverse missing rates. The code of the GT-TDI model will be available at https://github.com/KP-Zhang/GT-TDI.

## I. INTRODUCTION

COMPLETE traffic data plays an essential role in the applications of Intelligent Transportation Systems (ITS). As a set of sequential observations, traffic data can be gathered by different types of sensors, including stationary sensors (e.g., loop detectors) and mobile sensors (e.g., GPS probes). However, data missing is a common problem for both stationary and mobile sensors. Stationary sensors may produce faulty readings due to various causes such as malfunctioning hardware, power outages, and transmission errors, meanwhile mobile sensors are often complained about data sparseness with highly erratic spatial and temporal resolutions [1, 2]. Missing data problems will cast a shadow over the real-time monitoring of traffic conditions and hinder downstream applications of ITS from providing reliable traffic information. In this context, traffic data imputation becomes a necessity to address the problem of missing data.

Incomplete traffic data from a large road network are multidimensional time series that have different missing patterns under various missing rates. In general, the missing patterns are categorized as random missing (RM) and non-random missing (NM) [3, 4]. In a random missing pattern, the data points are lost without regularity, which may result from transmission packet loss or power network failure. In a non-random missing pattern, continuous missing and correlation corruption occur during a certain period, which may be the consequence of sensor malfunctioning, or regular maintenance [4]. Fig. 1 presents intuitive examples of two missing patterns at a sensor level and a network level. Compared with random missing patterns, non-random missing is trickier to cope with since that data may be lost for several hours or days over sensors in a road network. The long-term data missing will bring turmoil to spatiotemporal correlations and consequently hurt the imputation performance.

The key to imputing traffic data is to retrieve spatiotemporal information from observed data to help imputation models capture spatiotemporal correlations. Recently, a wide range of methods has been employed to make an accurate imputation [3, 5-8]. These methods tried to solve the fundamental problem of capturing spatiotemporal corrections by utilizing either time series modeling [9, 10] or matrix/tensor factorization [11, 12]. Since the corrupted data will undermine the spatiotemporal characteristics, it is not a trivial problem to capture spatiotemporal correlations from large-scale traffic data, especially when the missing rate is extremely large.

In this context, we propose a Graph Transformer for Traffic Data Imputation (GT-TDI) model to impute missing data. Specifically, a novel semantic description is introduced to store network-wide spatial and temporal information of traffic data, which can significantly improve the capability

The research is funded by the National Natural Science Foundation of China (62002101, 71772195), the Science and Technology Planning Project of Henan (222102210143), and the Fundamental Research Funds for the Henan Provincial Colleges and Universities (2018RCJH16). (*Corresponding author: Na Xie; Zhengbing He*)
K. Zhang is with the College of Electrical Engineering, Henan University of Technology, Zhengzhou 450001, China and the Department of Automation, Tsinghua University, Beijing 100084.
L. Wu is with the College of Electrical Engineering, Henan University of Technology, Zhengzhou 450001, China.
L. Zheng is with the School of Traffic and Transportation Engineering, Central South University, Changsha 410083, China.
N. Xie is with the School of Management Science and Engineering, Central University of Finance and Economics, Beijing 100081, China. (e-mail: xiena@cufe.edu.cn)
Z. He is with the Beijing Key Laboratory of Traffic Engineering, Beijing University of Technology, Beijing 100124, China. (e-mail: he.zb@hotmail.com)

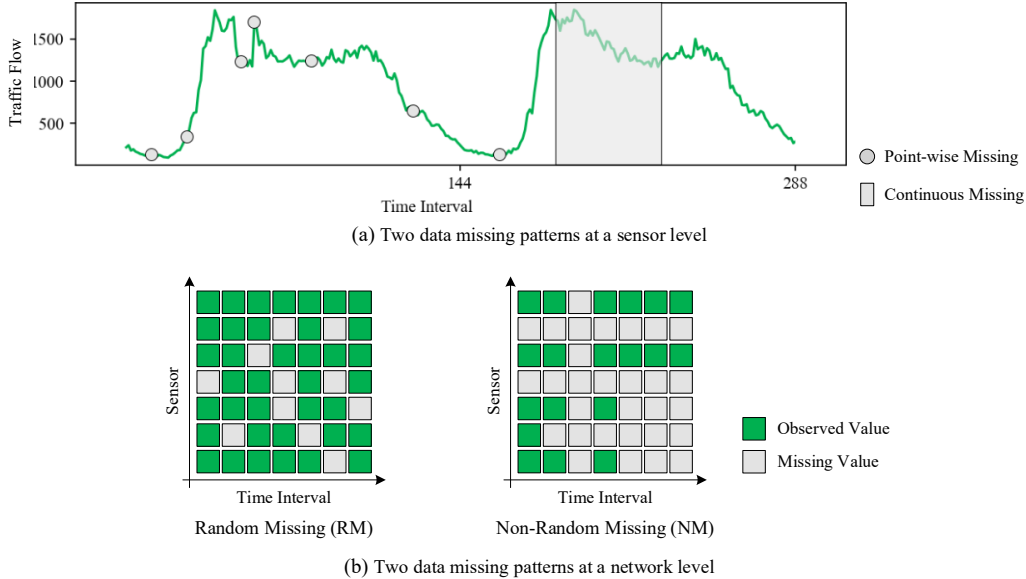

Fig. 1. Two data missing patterns.

of the GT-TDI model to capture spatiotemporal correlations at a network level. In the GT-TDI model, a road network is treated as a graph, where various sensors are denoted as nodes and their social connectivity is represented by edges. The graph will be processed by Graph Neural Networks (GNN) [13]. Then, the semantic description shared by the observed data and incomplete data is encoded by a word2vec model [14] before feeding into the residual network of the GT-TDI model. Finally, a Transformer encoder [15] is employed to carry out the imputation task.

Notably, the proposed GT-TDI model treats a sequence of observations per day from a sensor as a sentence of a corpus in Natural Language Processing (NLP). To take into account the intra-day trend of traffic states and maintain a proper granularity [16, 17], each sequence of observations is further divided into several slices, which can be recognized as words in a sentence. To store spatial and temporal information of each slice/word at a road network level, this study introduces the semantic description, which describes slices verbally. The semantic description consists of various labels, including the road ID, sensor ID, sensor position, flow direction, month, day, day of week, and slice ID, which are shared by observed traffic data and incomplete data. Then, the problem of traffic data imputation is converted to the problem of restoring the incomplete slices/words by mapping them into the traffic dataset/corpus with the help of the shared semantic description. Differently, in existing Transformer-based NLP models, the positional encoding only provides local information of each word in a sentence; while in this paper, the semantic description contains global spatial and temporal information of each slice in the road network.

The merits of the semantic description have two folds. (1) The semantic description includes spatial and temporal information with different granularity, which helps the imputation model understand the spatiotemporal semantic of the road network. (2) The global spatial and temporal information in the semantic description allows useful knowledge to be refined from one slice of traffic data to another at a network level. The abovementioned merits guarantee the imputation model precisely captures spatiotemporal correlations to improve the imputation performance. The main contributions of this study are summarized as follows:

- By introducing the semantic description, we provide new insight into the traffic data imputation, in which the spatiotemporal information of traffic data can be effectively described as a flexible semantic description to help the imputation model capture spatiotemporal correlations with proper granularity.
- To the best of our knowledge, the proposed GT-TDI model is the first successful attempt to impute traffic data in large-scale networks and extreme missing scenarios by utilizing Transformer.
- Based on real-world large-scale traffic data, we conduct extensive imputation experiments. The outcomes show that the proposed GT-TDI model can provide a much more accurate imputation than state-of-the-art models.

The remainder of the paper is organized as follows. Section II reviews the related literature on traffic data imputation and the applications of Transformer in traffic time series analysis. The methodology framework of the proposed GT-TDI model is proposed, and its details are described in Section III. Numerical experiments are conducted in Section IV. Finally, Section V draws some interesting conclusions.

## II. LITERATURE REVIEW

### A. Traffic Data Imputation Methods

Under diverse missing patterns, existing studies have proposed various imputation methods to cope with different types of traffic data (e.g., traffic flow, traffic speed, etc.). Among these methods, the Historical Average (HA) technique imputed missing values by simply averaging the historical values for given time intervals [18]. As a widely used model in time series data, the Autoregressive Integrated Moving Average (ARIMA) was employed to fill different types of missing traffic data [5, 19]. Steimetz and Brownstone [20] proposed a typical Markov Chain Monte Carlo (MCMC)-based model to impute key observations in estimating the value of time savings of commuters by treating a missed value as a parameter of the model. Ni and Leonard [21] utilized a Bayesian network to approximate probability distributions of observed data and then imputed missing data with the help of MCMC. The Principal Component Analysis (PCA) techniques were used in many imputation problems [22, 23], in which the PCA helped to remove the relatively trivial details from the observed traffic data. To impute missing data, early methods mainly focus on capturing temporal correlations of traffic data, while barely taking their spatial information into account.

To overcome the abovementioned drawback, some studies attempt to consider both temporal and spatial information of traffic data when imputing missing values. Zhang and Liu [24] proposed Least Squares Support Vector Machines (LS-SVMs) to impute missing values of traffic flow data by applying spatiotemporal analysis over urban arterial streets. Tak et al. [25] considered spatial and temporal aspects of missing data when imputing traffic speed with a modified K-Nearest Neighbor (KNN) method. Laña et al. [26] proposed a spatial context sensing model to retrieve useful information from surrounding sensors and an automated clustering analysis tool to impute traffic data with the help of optimal pattern clusters. Li et al. [6] presented a hybrid spatiotemporal method for traffic data imputation, in which a prophet model was utilized to reflect temporal properties of traffic data, and an iterative random forest model was employed to consider spatial dependencies. Kaur et al. [27] introduced a statistically principled imputation framework to impute missing loop detector data with the help of neighboring detectors. With the consideration of spatial correlations, these studies demonstrate that traffic data imputation can benefit from neighboring sensors. Most of them mainly consider local spatial information from nearby locations, while global spatiotemporal correlations at a network level are not fully explored to further improve the performance.

Recently, matrix/tensor factorization methods have been widely studied to impute traffic data by folding a traffic data vector from a sensor to a multi-dimensional matrix. These methods make use of the low-rank property of spatiotemporal traffic data to restore a complete matrix for missing data. Zhu et al. [8] proposed a compressive sensing based algorithm to solve the missing data problem by converting traffic data imputation to a matrix completion problem. Ran et al. [28] introduced a low-n-rank tensor completion algorithm to impute missing data, which reconstructed traffic flow data in a four-way tensor pattern. Chen et al. [29] constructed spatiotemporal traffic data to a third-order tensor structure and employed Bayesian probabilistic matrix factorization to recover missing data. Chen et al. [3] proposed a scalable tensor learning model named Low-Tubal-Rank Smoothing Tensor Completion (LSTC-Tubal) to impute large-scale traffic data. Based on the joint matrix factorization, Jia et al. [30] introduced an imputation model to impute traffic congestion data, which can consider network-wide spatiotemporal information. Tensor factorization models perform well in imputing large-scale traffic data due to the capability of capturing spatiotemporal correlations from multi-dimensional data. However, tensor factorization requires traffic data with a low rank and has to be learned from scratch for new batches of missing data [31]. In addition, to impute traffic data with complex missing patterns and large missing rates, tensor factorization models might face challenges to effectively retrieve traffic features and provide robust imputation results because of the nonlinearity and complexity of spatiotemporal correlations [2].

Moreover, due to the strong capability of digging spatiotemporal correlations, various deep learning-based methods have also been proposed for traffic data imputation. Compared with tensor factorization methods, deep learning models can learn useful features directly from traffic data without additional assumptions and do not have to be trained for every new batch of missing data. Specifically, Duan et al. [32] presented a denoising stacked autoencoder (DSAE) model to carry out traffic-flow data imputation with the consideration of spatiotemporal correlations. Li et al. [33] proposed a Multi-View Learning Method (MVLM) to impute traffic-related time series data by combining Long Short-Term Memory (LSTM), Support Vector Regression (SVR), and collaborative filtering techniques to capture spatiotemporal correlations. Asadi and Regan [34] introduced an autoencoder framework by combining Convolutional Neural Networks (CNN) and bidirectional-LSTM to capture spatial and temporal patterns in traffic-flow data imputation. Yang et al. [35] applied a Spatio-Temporal Learnable Bidirectional Attention Generative Adversarial Networks (ST-LBAGAN) to impute missing traffic data for a large-scale road network. Zhang et al. [36]

proposed a Travel Times Imputation Generative Adversarial Network (TTI-GAN) for travel times imputation with the consideration of network-wide spatiotemporal correlations. Recently, GNNs have been increasingly applied in traffic data imputation by converting a road network to a graph structure, in which the nodes represent sensors and edges denote social connectivity among sensors with the help of a designed adjacency matrix. Wu et al. [37] employed an Inductive Graph Neural Network Kriging (IGNNK) model to recover traffic data for unsampled sensors on a road network with a predefined adjacency matrix to describe social connectivity. Zhang et al. [38] proposed a Graph Convolutional Bidirectional Recurrent Neural Network (GCBRNN) to impute and predict large-scale traffic data. Ye et al. [39] applied an encoder-decoder structure with Graph Attention Convolutional Networks (GACN) to impute traffic data, in which graph attention mechanism was utilized to learn spatial correlations and temporal convolutional layers were stacked to take into account temporal correlations. Liang et al. [2] proposed a Dynamic Spatiotemporal Graph Convolutional Neural Network (DSTGCN) to impute missing traffic data with the consideration of spatiotemporal correlations at a network level. Different from other GNN-based imputation models, the DSTGCN employed a graph structure estimation technique to provide a dynamic adjacency matrix, which was helpful to capture spatial correlations changing over time. For existing GNN-based models, the capability of capturing network-wide spatiotemporal corrections mainly depends on a fine-designed model structure, and a qualified representation of social connectivity of sensors, which may be insufficient to impute large-scale traffic data with complex missing patterns and diverse missing rates since no perfect rule of thumb to design deep learning models and represent social connectivity at this stage.

*B. Transformer in Traffic Time Series Analysis*

In the last few years, Recurrent Neural Network (RNN) and its variants (e.g., LSTM and Gated Recurrent Units (GRU)) have been widely used in traffic time series analysis (e.g., traffic state prediction, passenger demand prediction, etc.) [40, 41]. However, RNN-based models are difficult to train due to gradient vanishing and exploding problems, which results in the inefficiency of capturing long-term dependencies [42]. Therefore, a more resilient technique is required to effectively capture long-term correlations from large-scale time series data. As a dominant technique in NLP, Transformer can overcome the drawbacks of RNN and its variants with the help of an attention mechanism. This inspires many studies to carry out traffic time series analysis with Transformer-based models. Specifically, Lim et al. [43] proposed a Temporal Fusion Transformer (TFT) by combining high-performance multi-horizon forecasting with interpretable insights into temporal dynamics, which provided excellent performance in traffic state and other time series predictions. Grigsby et al. [44] proposed a long-range Transformer for multivariate time series prediction, which demonstrated a desirable performance on traffic state prediction. Based on a pre-trained Bidirectional Encoder Representations from Transformers (BERT) [45], Jin et al. [46] proposed TrafficBERT to predict long-range traffic flow by utilizing multi-head self-attention to capture spatiotemporal correlations. Ye et al. [47] presented a Meta Graph Transformer (MGT) to model the spatial and temporal dynamics of traffic states when predicting metro crowd flow and highway traffic flow. Reza et al. [48] presented a multi-head attention based Transformer model to predict traffic flow on the PeMS dataset. To predict traffic demand, Huang et al. [49] proposed a Dynamical Spatial-Temporal Graph Neural Network (DSTGNN) model by integrating a diffusion CNN and a modified Transformer. Based on self-attention from Transformer, Liu et al. [50] presented a prediction model for short-term traffic flow with the consideration of spatiotemporal corrections. Su et al. [51] proposed a Graph Diffusing Transformer (GDFormer) to predict traffic flow on the PeMS dataset, in which a Graph Diffusing Attention (GDA) module was designed to reflect the changing traffic flow among sensors. With the consideration of network-wide spatiotemporal corrections, Zheng et al. [52] presented a self-attention Graph Convolutional Network with Spatial, Sub-spatial, and Temporal blocks (SAGCN-SST) model to predict traffic speed on a large-scale road network. Zhang et al. [53] proposed a Graph Attention Transformer (Gatformer) to predict trajectories for autonomous vehicles by integrating Graph Attention Networks (GAT) [54] and Transformer. These Transformer-based models require completed data to keep functioning properly, which cannot be applied to the problem of traffic data imputation.

III. TRAFFIC DATA IMPUTATION WITH GRAPH TRANSFORMER

*A. The Flowchart of Our Model*

In this study, a traffic network is defined as a weighted directed graph $G = (V, E)$, where $V$ is a set of connected nodes (i.e., sensor) in the graph and $E$ is a set of edges indicating the social connectivity among nodes. Fig. 2 gives the flowchart of the GT-TDI model. Based on the raw data from sensors, the road structure and observed traffic data are extracted to prepare the input for the proposed model.

In reality, although some roads are directed, this study considers the graph $G$ as an undirected graph to take into account the bidirectional impact of traffic congestion or failure, which will be bidirectionally propagated to upstream and downstream roads [55]. Then, based on the road structure, geography edges are retrieved for the graph $G$.

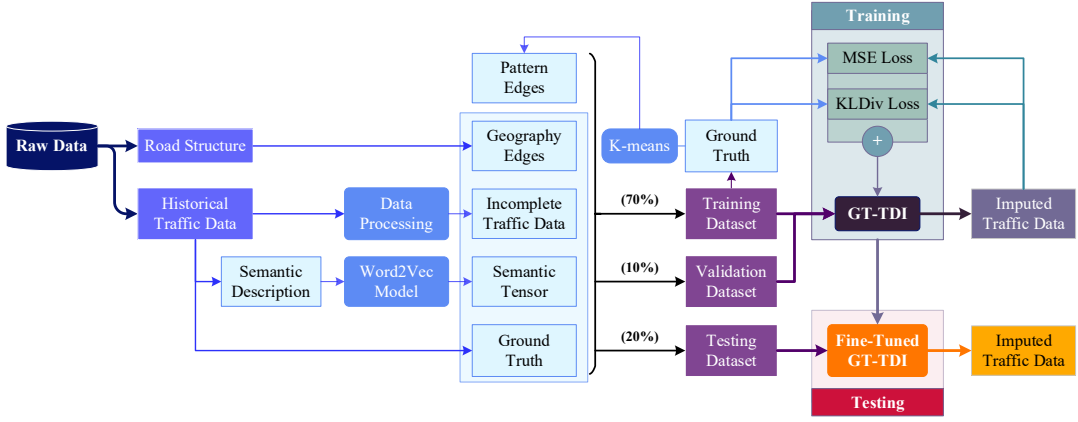

Fig. 2. The flowchart of the GT-TDI model.

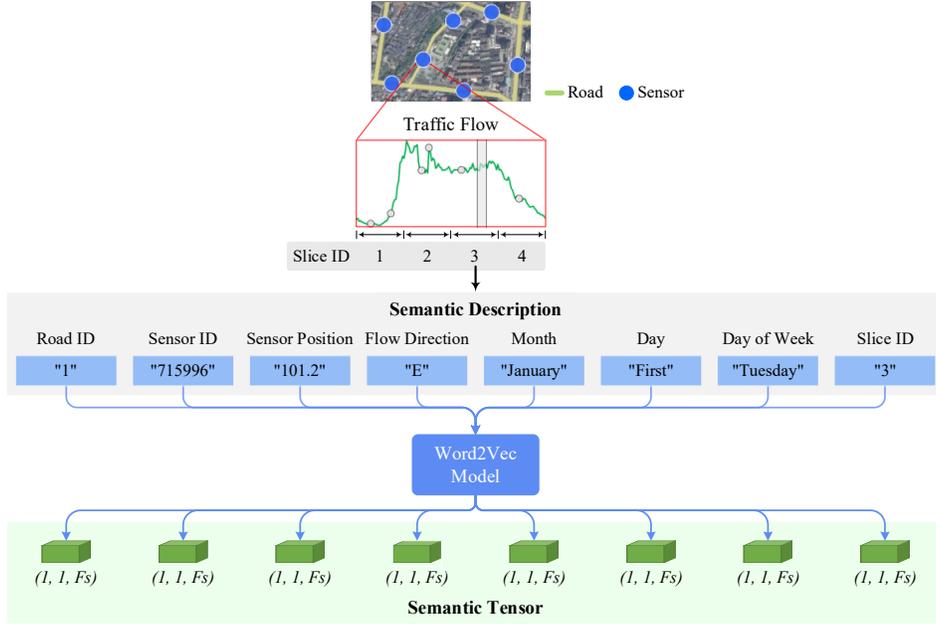

Fig. 3. Semantic description.

With the historical traffic data, incomplete traffic data are generated following the data processing method applied by Chen et al. [4]. The shared semantic description of the observed traffic data and the incomplete data is also prepared. The details of the semantic description can be found in Fig. 3, where the traffic flow from Sensor 715996 on January first is taken as an example. It can be found that the semantic description has eight labels that contain spatial and temporal information of Slice 3 in the road network. Then, the semantic description is fed into a word2vec model to generate a semantic tensor with a dimension of $(1, 1, Fs)$ for each label of the slice, where the first dimension is the sensor number, the second dimension is the slice number, and $Fs$ is the vector size of the word2vec model. The ground truth of the historical data is also retrieved to train, validate, and test the proposed model. The geography edges, incomplete traffic data, semantic tensor, and ground truth are separated into the training dataset, validation dataset, and testing dataset with a ratio of 70%, 10%, and 20%, respectively.

Moreover, to account for the similarity of the temporal patterns of traffic states across a road network, based on the ground truth data from the training dataset, pattern edges are retrieved with the help of K-means. In this study, 5 neighbors are selected for each sensor according to their traffic state similarities. Thus, the edge $E$ consists of two components: geography edges and pattern edges. The pattern edges then join the three datasets (i.e., training dataset, validation dataset, and testing dataset) to train, validate and test the GT-TDI model. The training dataset and the validation dataset are utilized to tune and validate the GT-TDI model with two loss functions (i.e., Mean squared error (MSE) loss and Kullback-Leibler divergence (KL Div) loss). The fine-tuned GT-TDI model is tested on the testing dataset to provide imputed traffic data.

*B. Model Input and Output*

As shown in Fig. 4, the incomplete traffic data $Y_I$, edge $E$, and semantic tensor $P$ are taken as the input of the GT-TDI model. The incomplete traffic data $Y_I$ has a dimension of *(D, S, N)*, where $D$ represents the number of days of collecting data, $S$ represents the sensor number of the road network, and $N$ is the number of observations during a day. To take into account the intra-day trend of traffic states, the traffic data of each day of a sensor are further divided into $M$ slices with a slice ID $m$ ($m=1, 2,..., M$) for each slice. Then the dimension of $Y_I$ becomes *(D∗M, S, N/M)*. The Edge $E$ has a dimension of *(2, W))*, where 2 represents two connected sensors, and $W$ is the number of edges. The semantic tensor $P$ has a dimension of *(D∗M, S, F')*, where $F'$ equals the product of the label number and $F_S$.

As a data imputation model, the GT-TDI model outputs imputed traffic data $\hat{Y}$ with a dimension of *(D∗M, S, N/M)*. The output can be written as

$$Output[\hat{Y}] = \text{GT-TDI}(Input[Y_I, E, P]) \quad (1)$$

*C. Architecture of the GT-TDI Model*

The architecture of the GT-TDI model is demonstrated in Fig. 4, which consists of three modules: the graph module, the semantic module, and the Transformer module.

In the graph module, the incomplete traffic data $Y_I$ and corresponding edge $E$ are fed into a Graph Transformer [56], which is a GNN with multi-head attention from Transformer. In the Graph Transformer shown in Fig. 5, multi-head attention is employed to execute graph learning with edge features. Specifically, given node features $H^{(l)} = \{h_1^{(l)}, h_2^{(l)}, ..., h_n^{(l)}\}$, the Graph Transformer calculates multi-head attention for each edge from $j$ to $i$ as follows:

$$q_{c,i}^{(l)} = W_{c,q}^{(l)} h_i^{(l)} + b_{c,q}^{(l)} \quad (2)$$

$$k_{c,j}^{(l)} = W_{c,k}^{(l)} h_j^{(l)} + b_{c,k}^{(l)} \quad (3)$$

$$e_{c,ij} = W_{c,e} e_{ij} + b_{c,e} \quad (4)$$

$$\alpha_{c,ij}^{(l)} = \frac{\langle q_{c,i}^{(l)}, k_{c,j}^{(l)} + e_{c,ij} \rangle}{\sum_{u \in \mathcal{N}(i)} \langle q_{c,i}^{(l)}, k_{c,u}^{(l)} + e_{c,iu} \rangle} \quad (5)$$

where $\langle q, k \rangle = \exp\left(\frac{q^T k}{\sqrt{d}}\right)$ is exponential scale dot-product function and $d$ is the hidden size of each head. For the $c^{th}$ head attention, the source feature $h_i^{(l)}$ and distant feature $h_j^{(l)}$ are transformed into query vector $q_{c,i}^{(l)} \in \mathbb{R}^d$ and key vector $k_{c,i}^{(l)} \in \mathbb{R}^d$, respectively using different trainable parameters $W_{c,q}^{(l)}, W_{c,k}^{(l)}, b_{c,q}^{(l)}, b_{c,k}^{(l)}$. The provided edge feature $e_{ij}$ will be encoded and added into the key vector as additional information for each layer.

After calculating the graph multi-head attention, a message aggregation is introduced from the distance $j$ to the source $i$:

$$v_{c,j}^{(l)} = W_{c,v}^{(l)} h_j^{(l)} + b_{c,v}^{(l)} \quad (6)$$

$$\hat{h}_i^{(l+1)} = \|_{c=1}^{C} \left[\sum_{j \in \mathcal{N}(i)} \alpha_{c,ij}^{(l)} (v_{c,j}^{(l)} + e_{c,ij})\right] \quad (7)$$

where the $\|$ is the concatenation operation for $C$ head attention. The multi-head attention matrix replaces the original normalized adjacency matrix of GCN as a transition matrix for message passing. The distance feature $h_j$ is transformed to $v_{c,j} \in \mathbb{R}^d$ for weighted sum.

Similar to GAT, if the Graph Transformer is employed as the last output layer, the graph module will average the multi-head outputs and remove the non-linear transformation as follows:

$$\hat{h}_i^{(l+1)} = \frac{1}{C} \sum_{c=1}^{C} \left[\sum_{j \in \mathcal{N}(i)} \alpha_{c,ij}^{(l)} (v_{c,j}^{(l)} + e_{c,ij}^{(l)})\right] \quad (8)$$

$$h_i^{(l+1)} = (1 - \beta_i^{(l)}) \hat{h}_i^{(l+1)} + \beta_i^{(l)} r_i^{(l)} \quad (9)$$

To improve the capability of graph learning, the graph module stacks two Graph Transformer layers with a LeakyReLU activation function in between.

As shown in Fig. 4, the CNN-based semantic module consists of two Conv1d layers with a LeakyReLU activation function and a residual connection with dropout to reduce overfitting and improve generalization. The semantic module takes the semantic tensor $P$ as the input to encode the semantic descriptions of incomplete traffic data. Then, the outputs from both the graph module (i.e., $\hat{Y}_G$ with a dimension of ($D*M, S, H_G$)) and the semantic module (i.e., $\hat{Y}_P$ with a dimension of ($D*M, S, H_P$)) are concatenated into a tensor $\hat{Y}_T$ with a dimension of ($D*M, S, H_T$), where $H_T = H_G + H_P$.

With the tensor $\hat{Y}_T$ as input, the Transformer module is introduced to carry out the imputation, which consists of a BatchNorm1D layer, a Transformer Encoder, two linear layers, and a LeakyReLU activation function. The BatchNorm1D layer is employed to accelerate and stabilize the training process of the GT-TDI model. Then, the Transformer Encoder is introduced to encoder the $\hat{Y}_T$ with multi-head attention. The detailed inner structure of the Transformer Encoder is demonstrated in Fig. 5. With the help of two linear layers and a LeakyReLU activation function, the Transformer Encoder outputs the imputed traffic data $\hat{Y}$.

The proposed GT-TDI model is trained end-to-end to minimize the difference between the ground truth traffic data and predicted data with two loss functions: MSE loss and KL Div loss. Only the distances of the missed traffic data and their corresponding imputed data are considered when calculating the MSE loss. The KL Div loss is introduced to measure the difference between the ground truth probability distribution of the historical traffic data and the probability distribution of the imputed traffic data. The optimizer Adam is used to optimize the GT-TDI model with a learning rate of 0.001, and the proposed model is implemented by PyTorch.

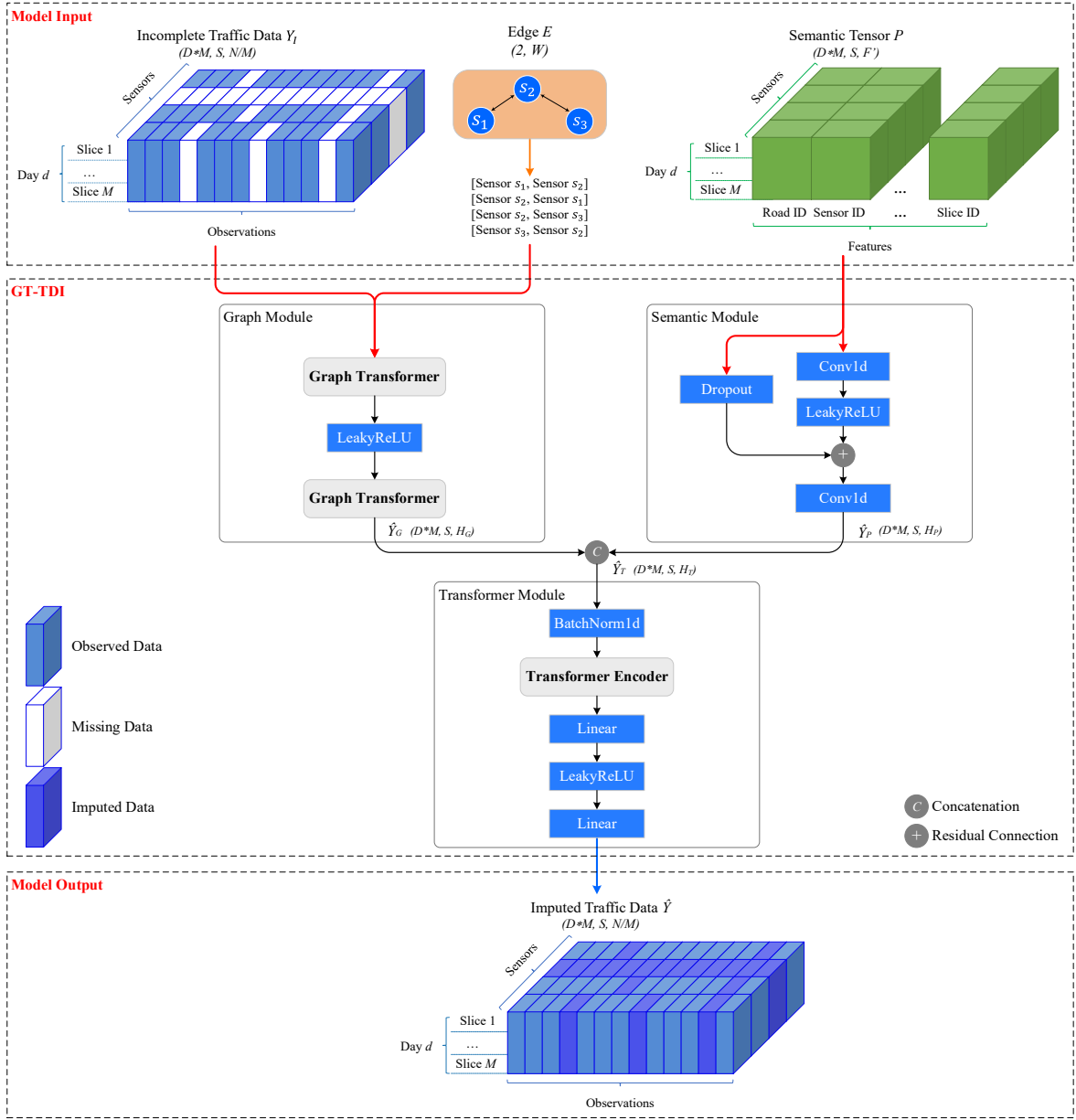

Fig. 4. The architecture of the GT-TDI model.

## IV. EXPERIMENTS AND RESULTS

### A. Study Site Selection

This study utilizes the traffic flow data from the PeMS to validate the performance of the proposed GT-TDI model. The traffic flow data from the PeMS are measured every 30s from over 15000 individual detectors deployed statewide in the freeway systems across California. The studied area is shown in Fig. 6 with red nodes representing data collection sensors. The collected data are aggregated at 10-min intervals from District 7 in 2013. This dataset contains traffic flow and traffic speed collected from 1740 traffic measurement sensors over 100 days, where the data from the first 70 days are treated as the training dataset, the data from the last 20 days are treated as the testing dataset, and the data from the remaining 10 days are treated as the validation dataset. For a random missing pattern and a non-random missing pattern, we train and test the proposed model respectively.

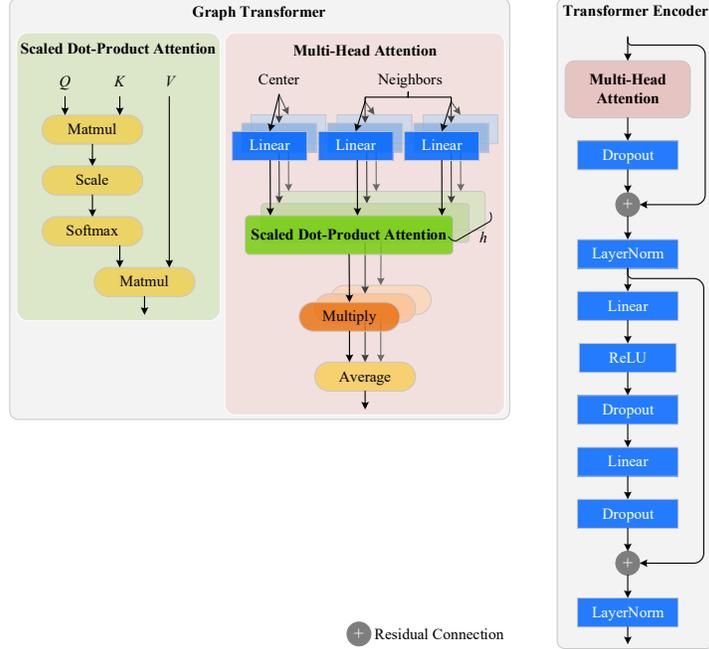

Fig. 5. The structure of Graph Transformer and Transformer Encoder.

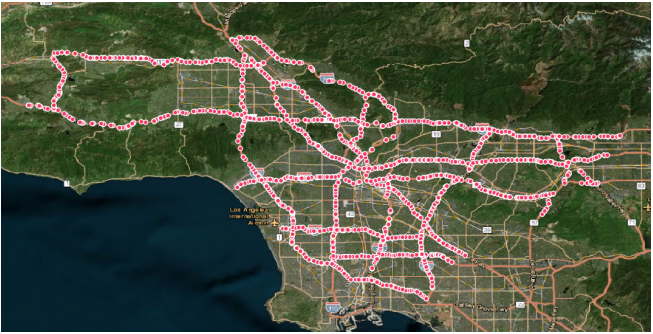

Fig. 6. The studied area.

This study utilizes the traffic flow data from the PeMS to validate the performance of the proposed GT-TDI model. The traffic flow data from the PeMS are measured every 30s from over 15000 individual detectors deployed statewide in the freeway systems across California. The studied area is shown in Fig. 6 with red nodes representing data collection sensors. The collected data are aggregated at 10-min intervals from District 7 in 2013. This dataset contains traffic flow and traffic speed collected from 1740 traffic measurement sensors over 100 days, where the data from the first 70 days are treated as the training dataset, the data from the last 20 days are treated as the testing dataset, and the data from the remaining 10 days are treated as the validation dataset. For a random missing pattern and a non-random missing pattern, we train and test the proposed model respectively.

This study utilizes the traffic flow data from the PeMS to validate the performance of the proposed GT-TDI model. The traffic flow data from the PeMS are measured every 30s from over 15000 individual detectors deployed statewide in the freeway systems across California. The studied area is shown in Fig. 6 with red nodes representing data collection sensors. The collected data are aggregated at 10-min intervals from District 7 in 2013. This dataset contains traffic flow and traffic speed collected from 1740 traffic measurement sensors over 100 days, where the data from the first 70 days are treated as the training dataset, the data from the last 20 days are treated as the testing dataset, and the data from the remaining 10 days are treated as the validation dataset. For a random missing pattern and a non-random missing pattern, we train and test the proposed model respectively.

The imputation performance is evaluated by two measurements: Mean Arctangent Absolute Percentage Error (MAAPE) [57] and Root Mean Squared Error (RMSE), formulated as

$$MAAPE = \frac{1}{N}\sum_{i=1}^{N}(arctan(|\frac{Y_i-\hat{Y}_i}{Y_i}|)) \quad (10)$$

$$RMSE = \sqrt{\frac{1}{N}\sum_{i=1}^{N}(Y_i - \hat{Y}_i)^2} \quad (11)$$

where $Y_i$ and $\hat{Y}_i$ are the $i^{th}$ ground truth and the imputed value of traffic data, respectively. This study utilizes MAAPE to avoid infinite or undefined values of Mean Absolute Percentage Error (MAPE) when the ground truth is zero or close to zero.

*B. Comparison Methods*

In the numerical experiment, we introduce conventional methods, tensor factorization methods, and deep learning-based methods to validate the effectiveness of the proposed GT-TDI.

Conventional methods include:

*KNN*: KNN is a well-known method in traffic data imputation. When imputing the missed data, K nearest data points are selected for the corrupted data, and then the missing entry can be imputed accordingly.

*PPCA*: Probabilistic Principal Component Analysis (PPCA) assumes a model where the observed data depends on the latent variables $Y = Wx + \mu + \varepsilon$. Y represents observed data, x denotes a latent variable from a Gaussian distribution, $\mu$ prevents the PPCA model from obtaining a zero mean, and $\varepsilon$ is isotropic noise from a Gaussian distribution. The PPCA method is inspired by [22].

Tensor factorization methods include:

*LSTC-Tubal*: As a type of tensor factorization method, the LSTC-Tubal is developed by Chen et al. [3]. The LSTC-Tubal introduces a scalable tensor nuclear norm minimization scheme by integrating linear unitary transformation, which organizes traffic data into a third-order tensor (location × time of day × day).

*LRTC-TNN*: The Low-Rank Tensor Completion with Truncated Nuclear Norm (LRTC-TNN) is proposed by Chen et al. [4]. To capture spatiotemporal correlations in traffic data, this method employs a universal rate parameter to control the degree of truncation on tensor modes.

Deep learning-based methods include:

*DAE*: The Denoising AutoEncoder (DAE) is similar to that proposed by [32, 58]. This deep learning-based model treats the incomplete vector as travel data that contain the observed normal data points and missing data points. This method transforms the traffic data imputation into the problem of incomplete data denoising.

*GCN-GRU*: Graph Convolutional Network with Gated Recurrent Units (GCN-GRU) is modified from the GT-TDI by replacing Graph Transformer with GCN and Transformer Encoder with GRU.

Our experiment platform is a server with a 3.2 GHz Intel Core i7 CPU, 64 GB memory, and two NVIDIA TITAN Xp graphics cards.

*C. Tests on Traffic Flow Data*

For a random missing pattern, Table 1 shows that the proposed GT-TDI model achieves the best performance under missing rates (MR) from 10% to 90%. Specifically, the KNN performs worst with MAAPE growing from 14.86 to 39.40 and RMSE increasing from 111.46 to 332.39, while the PPCA provides better performance. As for the tensor factorization methods, the LSTC and LRTC achieve impressive results with their MAAPEs raise from 6.85 to 29.60 and from 4.72 to 15.93, respectively as the MR grows. The deep learning-based counterparts (i.e., DAE and GCN-GRU) are better than the conventional methods (i.e., KNN and PPCA) while less appealing than the tensor factorization methods. The GT-TDI model outperforms the other counterparts with the MAAPE and RMSE growing from 4.53 to 6.52 and from 28.53 to 47.67, respectively. It is also worth noticing that the GT-TDI model can address the extreme case when the MR even reaches 90%. In addition, in three missing rates (i.e., 50%, 70%, and 90%), the imputed traffic speed data are compared with the ground truth in Fig. 7, which also shows the corresponding incomplete data. With the incomplete data as input, the proposed model is capable of accurately imputing the missed values even under the extreme missing rate (i.e., 90%).

For a non-random missing pattern, Table 2 shows that the proposed GT-TDI model achieves the best performance under MR from 10% to 90%. Specifically, the conventional methods (i.e., KNN and PPCA) provide undesirable performance under various MR. It can be found that the tensor factorization methods (i.e., LSTC-Tubal and LRTC-TNN) fail to maintain superiority under many missing rates compared with the conventional methods. As for the deep learning-based methods, the GCN-GRU model performs much better than the DAE model, whose MAAPE grows from 9.53 to 13.89 and RMSE increases from 66.57 to 79.95. In the non-random missing pattern, the GT-TDI model outperforms the other counterparts with the MAAPE and RMSE growing from 5.74 to 7.18 and from 37.22 to 54.62, respectively. It is also worth noticing that the GT-TDI model can address the extreme case when the MR reaches 90%. Fig. 8 shows the imputed traffic flow from the GT-TDI model under three missing rates (i.e., 50%, 70%, and 90%). It can be found that the imputed traffic flow data can precisely capture the main intra-day trend of the ground truth with extreme missing rates.

In summary, the conventional methods cannot accurately impute traffic flow due to their lack of considering the spatiotemporal correlations at a network level. Two tensor factorization methods can provide highly competitive performance in the random missing pattern while failing to maintain their momentums in the non-random pattern. Based on traditional deep learning methods, the results from the DAE and GCN-GRU are less appealing compared with the tensor factorization methods in the random missing pattern, although the GCN-GRU achieves plausible results in the non-random missing pattern. This may indicate that the deep learning methods may be ineffective to capture network-wide spatiotemporal corrections without spatiotemporal semantic understanding or well-designed models. The GT-TDI demonstrates that a proper network structure with Transformer and an informative semantic tensor from the shared semantic descriptions of the observed data and incomplete data can realize the maximum capability of a deep learning-based method when imputing large-scale traffic data.

Table 1 Comparison of performance (MAAPE/RMSE) by seven models for traffic flow imputation in a random missing pattern.

| MR | KNN | PPCA | LSTC-Tubal | LRTC-TNN | DAE | GCN-GRU | GT-TDI |
|---|---|---|---|---|---|---|---|
| 10% | 14.86/111.46 | 11.06/70.72 | 6.85/51.20 | 4.72/31.80 | 10.31/70.42 | 8.31/60.42 | **4.53/28.53** |
| 20% | 18.52/147.83 | 11.05/70.66 | 7.11/53.33 | 5.44/34.16 | 10.64/70.56 | 8.50/63.29 | **4.65/29.70** |
| 30% | 22.72/186.78 | 11.15/71.02 | 7.37/54.98 | 6.12/40.18 | 10.16/70.93 | 9.64/63.99 | **4.74/30.76** |
| 40% | 26.93/224.70 | 11.20/71.37 | 7.59/56.71 | 6.30/42.99 | 10.20/71.22 | 9.97/68.57 | **4.89/32.24** |
| 50% | 31.51/264.71 | 11.29/71.81 | 7.92/58.67 | 6.78/46.65 | 10.89/71.69 | 10.31/70.52 | **5.08/33.96** |
| 60% | 35.73/300.33 | 11.38/72.76 | 8.67/62.01 | 7.44/52.15 | 11.30/72.58 | 11.25/71.88 | **5.26/35.56** |
| 70% | 39.51/332.17 | 11.43/74.03 | 10.98/69.93 | 8.56/62.71 | 11.39/73.54 | 11.34/72.60 | **5.56/38.33** |
| 80% | 42.29/353.82 | 14.95/93.91 | 17.15/98.12 | 9.58/65.70 | 14.23/90.61 | 13.90/86.80 | **5.86/41.29** |
| 90% | 39.40/332.39 | 34.97/255.59 | 29.60/173.96 | 15.93/92.18 | 33.58/243.79 | 30.57/179.65 | **6.52/47.67** |

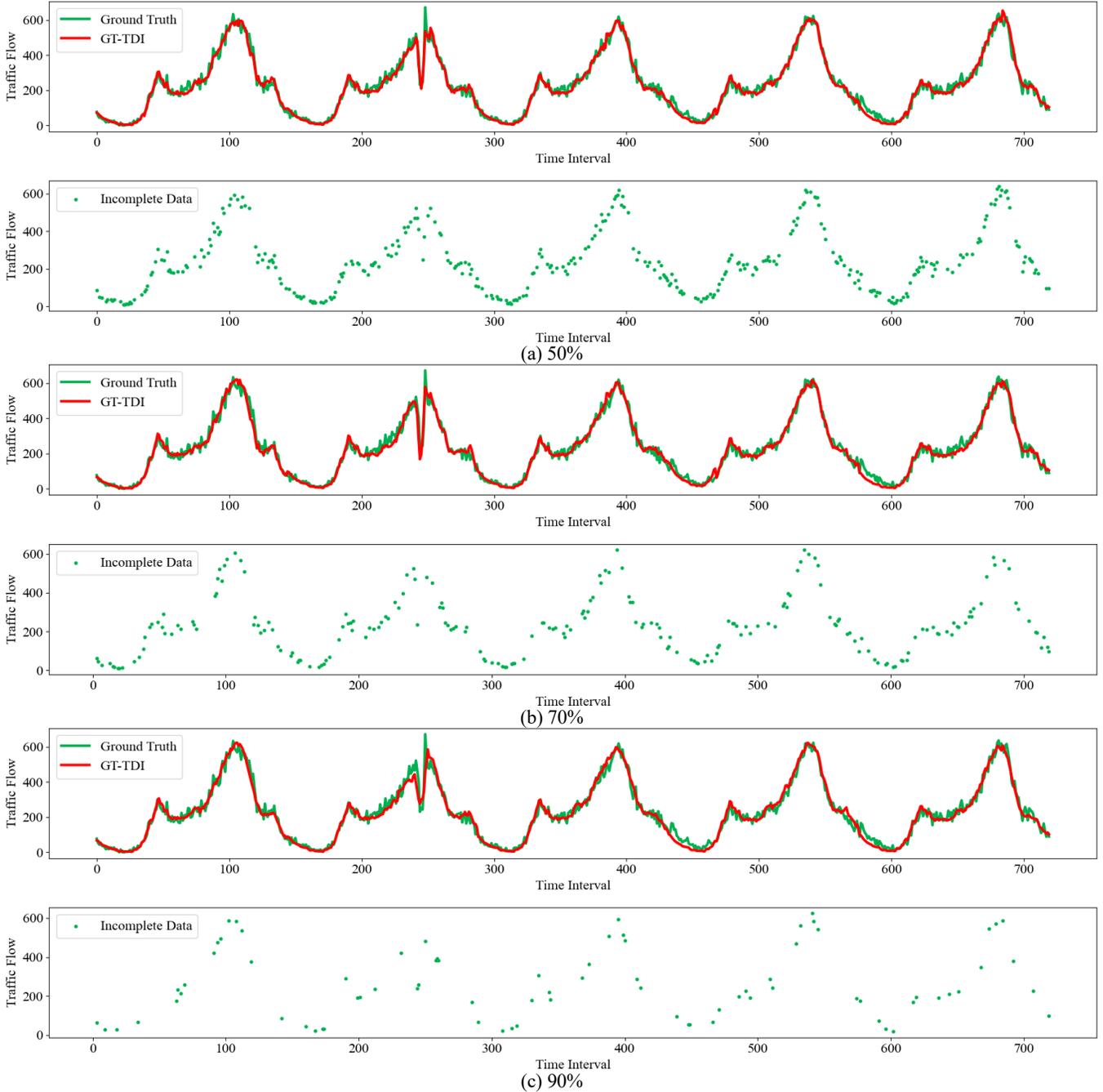

Fig. 7. Imputed traffic flow of the GT-TDI model under three missing rates in a random missing pattern.

Table 2 Comparison of performance (MAAPE/RMSE) by seven models for traffic flow imputation in a non-random missing pattern.

| MR | KNN | PPCA | LSTC-Tubal | LRTC-TNN | DAE | GCN-GRU | GT-TDI |
|---|---|---|---|---|---|---|---|
| 10% | 38.26/328.33 | 33.71/251.42 | 39.87/244.78 | 6.53/49.28 | 36.72/309.34 | 9.53/66.57 | **5.74/37.22** |
| 20% | 35.20/259.45 | 35.15/254.90 | 43.53/278.03 | 10.80/92.36 | 37.83/318.21 | 9.62/67.31 | **5.74/37.60** |
| 30% | 40.01/269.55 | 34.78/255.19 | 45.85/296.41 | 22.08/177.79 | 39.12/320.25 | 9.88/68.20 | **5.73/37.91** |
| 40% | 46.53/327.90 | 35.25/256.11 | 47.47/309.72 | 33.97/265.93 | 46.43/325.94 | 10.16/69.63 | **5.81/38.89** |
| 50% | 37.39/266.57 | 34.68/254.93 | 48.88/324.90 | 44.25/351.20 | 38.85/320.79 | 10.53/71.20 | **5.93/39.99** |
| 60% | 35.69/276.00 | 34.82/254.59 | 50.00/333.42 | 52.16/424.50 | 37.21/305.42 | 10.86/72.58 | **6.06/41.45** |
| 70% | 38.35/326.71 | 34.97/255.92 | 51.19/345.05 | 58.56/485.44 | 41.42/367.19 | 11.04/74.09 | **6.31/45.11** |
| 80% | 41.96/385.92 | 35.12/255.06 | 52.15/356.04 | 64.07/550.25 | 45.52/390.18 | 12.91/79.18 | **6.56/47.20** |
| 90% | 38.12/268.70 | 35.21/255.45 | 53.26/367.46 | 69.45/619.81 | 44.83/385.09 | 13.98/79.95 | **7.18/54.62** |

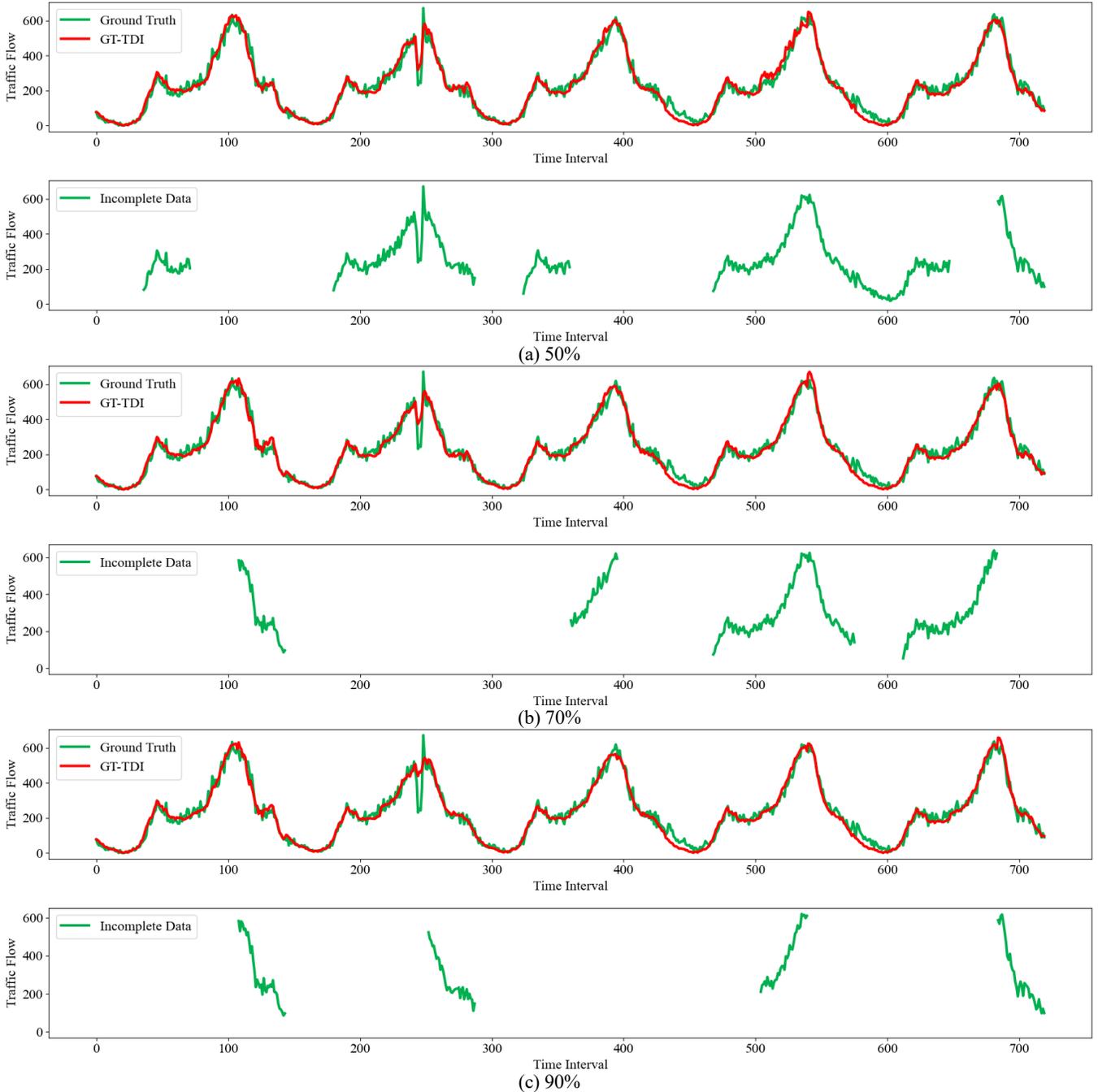

Fig. 8. Imputed traffic flow of the GT-TDI model under three missing rates in a non-random missing pattern.

Table 3 Comparison of performance (MAAPE/RMSE) by seven models for traffic speed imputation in a random missing pattern.

| MR | KNN | PPCA | LSTC-Tubal | LRTC-TNN | DAE | GCN-GRU | GT-TDI |
|---|---|---|---|---|---|---|---|
| 10% | 8.71/6.45 | 7.44/4.95 | 3.41/2.39 | 2.93/1.99 | 6.67/4.67 | 5.37/4.01 | **2.93/1.89** |
| 20% | 9.92/6.76 | 7.42/4.96 | 3.61/2.54 | 3.22/2.18 | 6.72/4.73 | 5.66/4.04 | **2.94/1.99** |
| 30% | 11.09/7.10 | 7.45/4.97 | 3.82/2.71 | 3.55/2.40 | 6.34/4.60 | 6.02/4.15 | **2.92/1.93** |
| 40% | 11.66/7.00 | 7.48/5.00 | 4.09/2.89 | 3.87/2.63 | 6.54/4.67 | 6.39/4.50 | **3.13/2.12** |
| 50% | 12.03/7.13 | 7.53/5.04 | 4.41/3.11 | 4.29/2.91 | 7.08/4.75 | 6.70/4.67 | **3.30/2.25** |
| 60% | 12.23/7.36 | 7.52/5.10 | 4.80/3.38 | 4.74/3.23 | 7.33/4.79 | 7.30/4.75 | **3.41/2.35** |
| 70% | 12.36/7.52 | 7.68/5.18 | 5.34/3.73 | 5.23/3.60 | 7.31/4.68 | 7.27/4.62 | **3.57/2.44** |
| 80% | 12.42/7.38 | 8.74/5.90 | 6.18/4.25 | 5.92/4.09 | 8.56/5.67 | 8.36/5.43 | **3.52/2.58** |
| 90% | 12.41/7.15 | 20.59/12.33 | 7.64/5.09 | 6.82/4.74 | 17.07/12.15 | 15.54/8.95 | **3.88/2.96** |

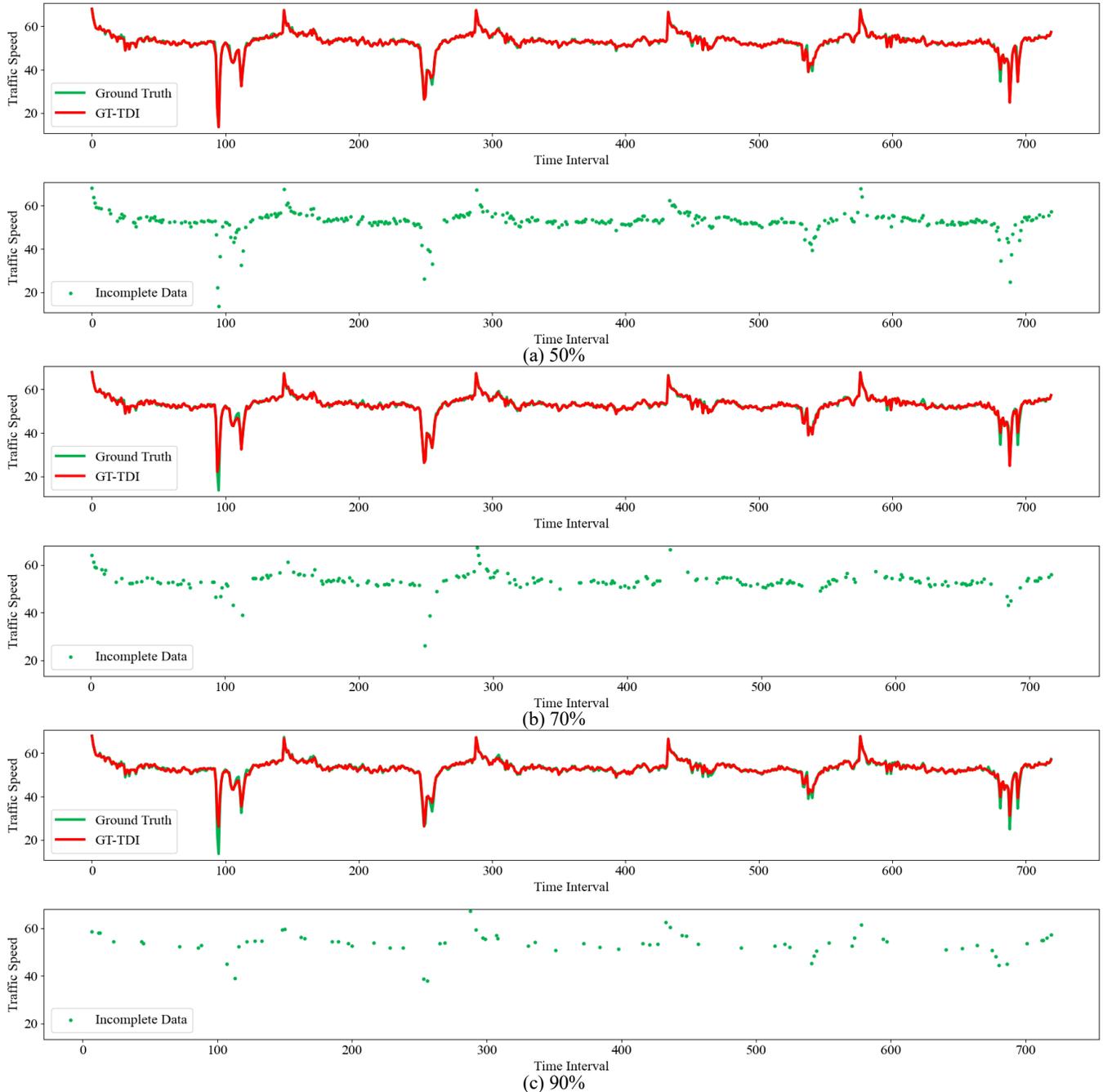

Fig. 9. Imputed traffic speed of the GT-TDI model under three missing rates in a random missing pattern.

**Table 4** Comparison of performance (MAAPE/RMSE) by seven models for traffic speed imputation in a non-random missing pattern.

| MR | KNN | PPCA | LSTC-Tubal | LRTC-TNN | DAE | GCN-GRU | GT-TDI |
|---|---|---|---|---|---|---|---|
| 10% | 9.47/6.25 | 8.60/5.67 | 9.17/5.96 | 7.05/4.69 | 7.53/5.02 | 7.22/4.81 | **6.81/4.41** |
| 20% | 9.56/6.32 | 8.63/5.70 | 9.50/6.08 | 7.41/4.90 | 8.22/5.44 | 7.38/4.87 | **6.95/4.51** |
| 30% | 9.77/6.49 | 8.98/5.96 | 9.71/6.25 | 7.57/5.08 | 9.38/6.30 | 7.50/5.04 | **7.07/4.66** |
| 40% | 9.88/6.67 | 9.21/6.21 | 9.95/6.39 | 7.92/5.37 | 10.06/6.82 | 7.86/5.33 | **7.10/4.58** |
| 50% | 10.11/6.84 | 9.62/6.50 | 10.37/6.64 | 8.25/5.63 | 11.09/7.57 | 8.21/5.60 | **7.17/4.74** |
| 60% | 11.88/8.15 | 9.88/6.77 | 10.86/6.90 | 8.66/5.95 | 12.59/8.65 | 8.76/6.02 | **7.31/4.90** |
| 70% | 12.63/9.01 | 10.04/7.16 | 11.65/7.39 | 9.17/6.54 | 15.92/11.35 | 9.25/6.76 | **7.54/5.14** |
| 80% | 15.95/13.27 | 11.95/9.94 | 13.17/8.21 | 10.32/8.61 | 19.92/16.62 | 11.64/9.71 | **7.33/5.28** |
| 90% | 19.11/13.40 | 13.00/9.11 | 16.05/10.00 | 78.54/55.15 | 20.03/14.06 | 18.48/12.98 | **7.95/6.52** |

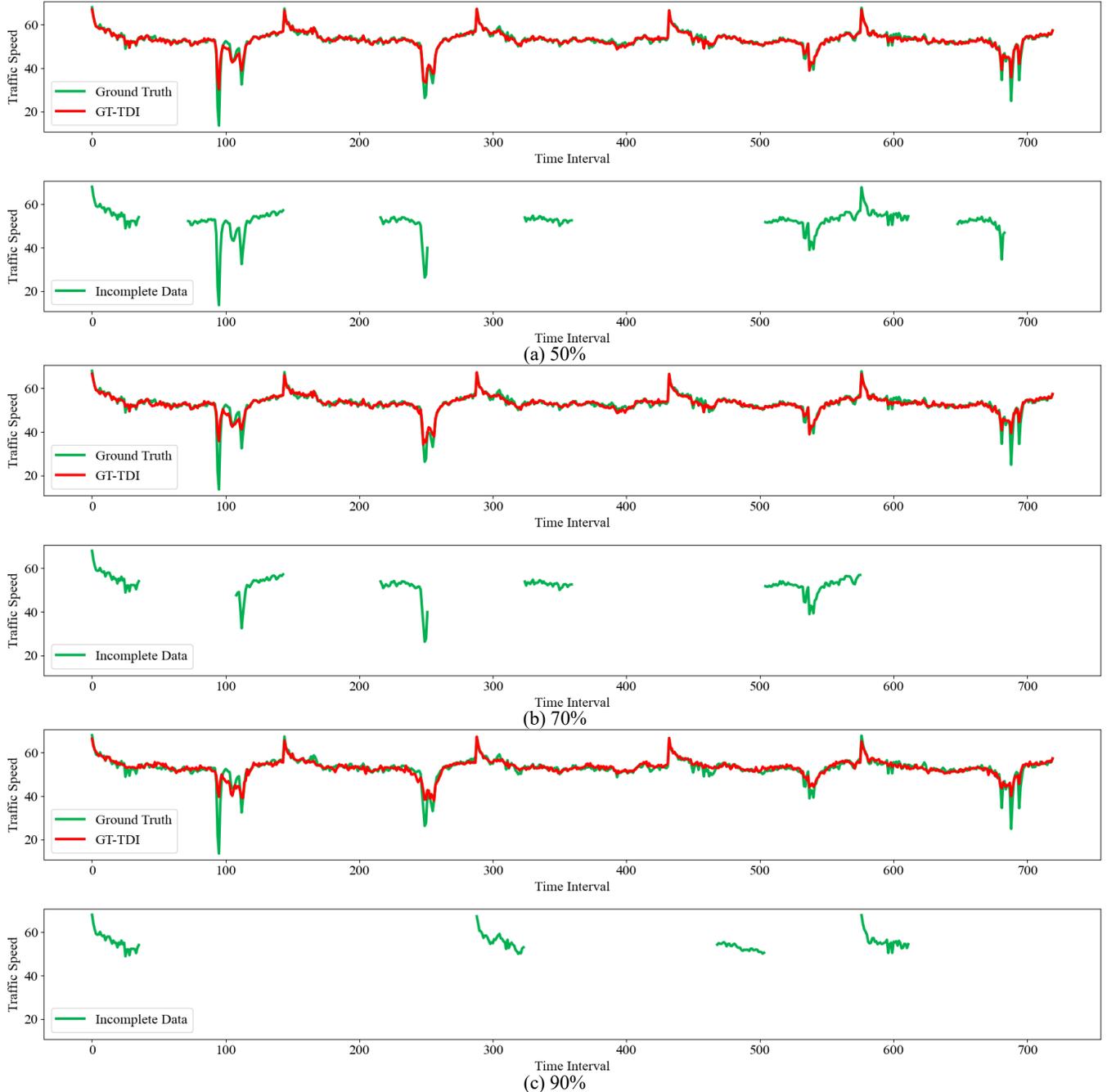

Fig. 10. Imputed traffic speed of the GT-TDI model under three missing rates in a non-random missing pattern.

*D. Tests on Traffic Speed Data*

Table 3 details the results of seven models when imputing traffic speed data in a random missing pattern. Obviously, the proposed GT-TDI model achieves the best performance under MR from 10% to 90%. The KNN performs worst with MAAPE growing from 8.71 to 12.42 and RMSE increasing from 6.45 to 7.38, while the PPCA provides better performance in most missing rates. As for the tensor factorization methods, the LSTC-Tubal and LRTC-TNN achieve competitive results with their MAAPEs raise from 3.41 to 7.64 and from 2.93 to 6.82, respectively as the MR grows. The deep learning-based methods (i.e., DAE and GCN-GRU) are better than the conventional methods (i.e., KNN and PPCA) while less appealing than the tensor factorization methods. Moreover, the GT-TDI model outperforms the other counterparts with the MAAPE and RMSE growing from 2.93 to 3.88 and from 1.89 to 2.96, respectively. In addition, in three missing rates (i.e., 50%, 70%, and 90%), the imputed traffic speed data are compared with the ground truth in Fig. 9, which also demonstrates the corresponding incomplete data. With incomplete data as input, the GT-TDI model is capable of capturing the intra-day trend of traffic speed and accurately imputing the missed values in various missing rates.

For a non-random missing pattern, Table 4 shows that the proposed GT-TDI model achieves the best performance with MR from 10% to 90%. Specifically, the KNN performs poorly with MAAPE growing from 9.47 to 19.11 and RMSE increasing from 6.25 to 13.40, while the tensor factorization models (i.e., LSTC-Tubal and LRTC-TNN) provide better performance. The LRTC-TNN model provides impressive results in MR from 10% to 80%, while performs worst in the missing rate of 90%. The deep learning-based models (i.e., DAE and GCN-GRU) have a better performance compared with the conventional methods in most cases. The GT-TDI model outperforms the other counterparts with the MAAPE and RMSE growing from 6.81 to 7.95 and from 4.41 to 6.52, respectively. It is also worth noticing that the GT-TDI model can address the extreme case when the MR even reaches 90%. Moreover, the imputed traffic speed values from the GT-TDI model are drawn in Fig. 10 with the corresponding ground truth and incomplete data. It can be observed that, even with few slices available, the proposed GT-TDI model can accurately capture the intra-day trend of the traffic speed and provide a reasonable imputation to the missed values.

In summary, without the consideration of network-wide spatiotemporal correlations, the conventional methods cannot accurately impute traffic speed, especially when the MR is large. Two tensor factorization methods can provide highly competitive performance, which shows their advantages in imputing large-scale traffic data. Based on traditional deep learning methods, the results from the DAE and GCN-GRU are less appealing compared with the tensor factorization methods in many missing rates. This indicates that the deep learning methods may be ineffective to capture network-wide spatiotemporal corrections by solely relying on incomplete data and incompetent models. The results from the GT-TDI model demonstrate that the semantic descriptions, GNNs, and Transformer are promising to impute large-scale traffic speed data.

*E. Ablation Studies*

In this subsection, four experiments are designed to test whether the GT-TDI model could work well with various semantic descriptions, what effect does the number of slices have on the model performance, whether the introduction of the KLDiv loss brings merits in training the GT-TDI model, and whether the introduction of K-means can improve the imputation performance. The abovementioned experiments are based on the traffic flow data under a random missing pattern.

**Ablation Experiment 1:** As mentioned in Section 3.1, the semantic description consists of various components (i.e., road ID, sensor ID, sensor position, flow direction, month, day, day of week, and slice ID). To study the effectiveness of each component, this experiment collects the changes in performance while adjusting the semantic description. As shown in Table 5, only one component will be adjusted each time, and then the performance change is measured by the average MAAPE over the missing rates from 10% to 90%. "✓" and "✗" indicate whether the corresponding component is involved or not in the semantic description.

From Table 5, without the semantic description, the GT-TDI model performs worst with an average MAAPE of 12.76. After introducing the road ID in the semantic description, the average MAAPE of the proposed model decreases by 7.52%. The sensor ID brings a more significant improvement to the model performance by 12.20%. The sensor position information could further improve the performance by 8.20%. By adding the flow direction to the semantic description, the average MAAPE reduces from 9.51 to 9.02. With the consideration of the month of a year, the performance of the GT-TDI model is further improved with the average MAAPE decreasing from 9.02 to 8.23. The information of the day of a month further enhances the performance of the GT-TDI model by 9.36%. Moreover, the performance of the proposed model improves by 14.21% after introducing the day of week to the semantic description. Finally, the slice ID brings a significant improvement to the GT-DTI model by 18.28%.

The semantic description is effective to provide spatiotemporal information to the GT-TDI model to capture spatiotemporal correlations when imputing missing traffic data. Specifically, by introducing the spatial information (i.e., road ID, sensor ID, and sensor position) to the semantic description, the GT-DTI can effectively consider spatial correlations of traffic flow with improved performance.

Table 5 Performance of the GT-TDI model with various semantic descriptions.

| Road ID | Sensor ID | Sensor Position | Flow Direction | Month | Day | Day of Week | Slice ID | Average MAAPE |
|---|---|---|---|---|---|---|---|---|
| × | × | × | × | × | × | × | × | 12.76 (—) |
| ✓ | × | × | × | × | × | × | × | 11.80 (↓7.52%) |
| ✓ | ✓ | × | × | × | × | × | × | 10.36 (↓12.20%) |
| ✓ | ✓ | ✓ | × | × | × | × | × | 9.51 (↓8.20%) |
| ✓ | ✓ | ✓ | ✓ | × | × | × | × | 9.02 (↓5.15%) |
| ✓ | ✓ | ✓ | ✓ | ✓ | × | × | × | 8.23 (↓8.76%) |
| ✓ | ✓ | ✓ | ✓ | ✓ | ✓ | × | × | 7.46 (↓9.36%) |
| ✓ | ✓ | ✓ | ✓ | ✓ | ✓ | ✓ | × | 6.40 (↓14.21%) |
| ✓ | ✓ | ✓ | ✓ | ✓ | ✓ | ✓ | ✓ | 5.23 (↓18.28%) |

The sensor ID could be more valuable information since it empowers the proposed model to distinguish different traffic data of various sensors. The flow direction can help the GT-DTI model capture the difference between the traffic states of different directions on a road. Moreover, the GT-DTI model can benefit from the temporal information encoded in month, day, and day of week. Particularly, the day of week can inform the proposed model to explore the weekly pattern of traffic states. In addition, the slice ID can enable the GT-DTI model to effectively consider the intra-day trend of traffic states to improve its performance.

In summary, the semantic description plays a vital role in capturing the spatiotemporal correlations when imputing traffic data with the GT-TDI model. In addition, as a general description of a road network, the utilization of the semantic description can be easily extended to other traffic-related estimation problems (e.g., traffic state estimation, passenger demand prediction, crash risk prediction, etc.).

**Ablation Experiment 2:** Different numbers of slices could result in various granularities for the intra-day trend of traffic states, which will have a different effect on the performance of the GT-DTI model. A set of values (i.e., 0, 2, 4, 6, 8, and 12) are selected as the number of slices to test the performance of the GT-DTI model. As shown in Fig. 11, the average MAAPE of the missing rate from 10% to 90% is recorded for each number of slices. It can be observed that as the number grows the average MAAPE first drops and then increases. When the number of slices equals 0, there is no intra-day trend considered when carrying out the imputation, which results in the worst performance. When the number of slices reaches 4, the GT-DTI model provides the best performance. In summary, the GT-DTI model can benefit from the consideration of the intra-day trend of traffic states. Meanwhile, the number of slices larger than 4 will diminish the performance since a tiny granularity of the intra-day trend may not reflect meaningful traffic states.

**Ablation Experiment 3:** To validate the effectiveness of the KLDiv loss, the average MAAPE of the GT-TDI with or without the KLDiv loss is recorded after each training epoch. From Fig. 12, it can be found that the GT-TDI provides a better result with the help of the KLDiv loss, which indicates that the KLDiv loss could be a valuable supplement to training the GT-TDI. This can be explained that the KLDiv loss between the probability distribution of the ground truth traffic flow and the probability distribution of the imputed traffic flow could provide useful information for the GT-TDI model to perform better.

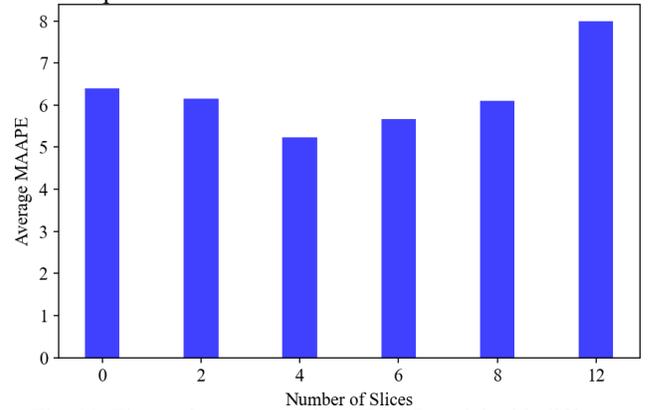
Fig. 11. The performance of the GT-TDI model with different numbers of slices.

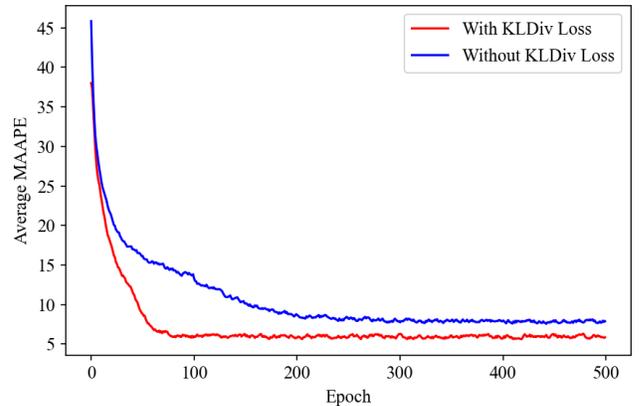
Fig. 12. The performance of the GT-TDI model with or without the KLDiv loss.

**Ablation Experiment 4:** In this study, K-means is introduced to provide pattern edges to the GT-TDI model to account for the similarity of the temporal patterns of the traffic states across the road network. The performance of the GT-TDI model without K-means is demonstrated in Fig.

13. It can be observed that the performance of the GT-TDI model degrades without K-means. This indicates that the performance of the GT-TDI model can benefit from the pattern edges provided by K-means.

In addition, the performance of the proposed model with different numbers of neighbors of K-means is drawn in Fig. 14. It can be found that the average MAAPE first drops as the number of neighbors increases from 1 to 5. Then, the average MAAPE gradually raises as the number of neighbors grows. The proposed GT-TDI model performs best when the neighbor number reaches 5. This can be explained that the proposed model can benefit from capturing temporal correlations from more neighbors. Meanwhile, a number of neighbors large than 5 will diminish the performance due to redundant features that may hurt the performance of the proposed GT-TDI model.

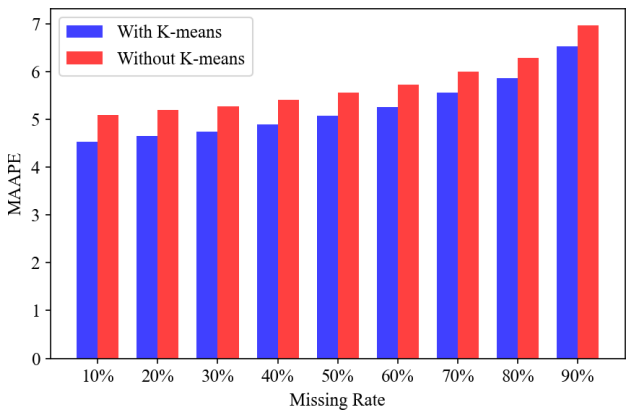

Fig. 13. The performance of the GT-TDI model with or without K-means.

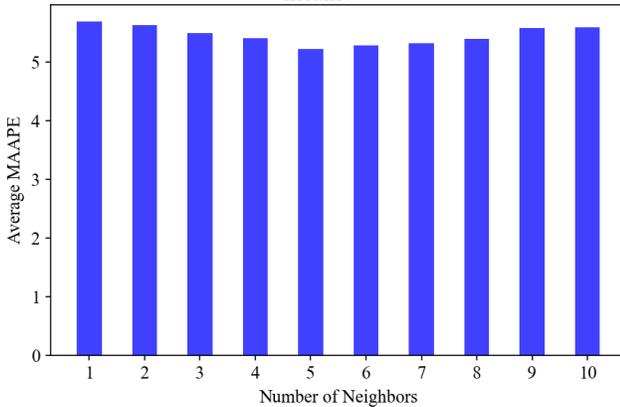

Fig. 14. The performance of the GT-TDI model with various numbers of neighbors.

## V. CONCLUSIONS

This study proposes a GT-TDI model to impute large-scale traffic data by combining GNNs, CNN-based residual networks, and Transformer. Meanwhile, the semantic description is introduced to help the GT-TDI model capture spatiotemporal correlations at a network level. The numerical experiments are carried out with the traffic data from the PeMS dataset. Compared with the conventional methods (i.e., KNN and PPCA), tensor factorization methods (i.e., LSTC-Tubal and LRTC-TNN), and deep learning-based methods (i.e., DAE and GCN-GRU), the GT-TDI model provides more accurate imputations, especially when the missing rate is large.

The contributions of this study are concluded as follows: (1) With the help of the semantic description, the GT-TDI model can well capture the spatiotemporal correlations for the traffic data imputation at a network level; (2) By introducing GNNs, and Transformer, the GT-TDI model can effectively impute traffic data in large-scale networks and extreme missing scenarios; (3) Based on the results from extensive imputation experiments, the GT-TDI model can provide much more accurate imputation than state-of-the-art models, especially under extreme missing scenarios.

Currently, the most suited number of slices requires manual selection. In the future, more advanced techniques (e.g., masked slices) will be explored to capture the intra-day trend of traffic states. Moreover, it could be more efficient if the GT-TDI model can directly retrieve spatiotemporal corrections from the semantic description without the help of word2vec models. In addition, following the success of the GT-TDI model in imputing large-scale traffic data, it is encouraging to extend the utilization of the semantic description to other large-scale traffic-related estimation problems (e.g., traffic state estimation, passenger demand prediction, crash risk prediction, etc.).

[42] S. Li, X. Jin, Y. Xuan, X. Zhou, W. Chen, Y.-X. Wang, and X. Yan, "Enhancing the locality and breaking the memory bottleneck of transformer on time series forecasting," *Advances in Neural Information Processing Systems,* vol. 32, 2019.

[43] B. Lim, S. Ö. Arık, N. Loeff, and T. Pfister, "Temporal fusion transformers for interpretable multi-horizon time series forecasting," *International Journal of Forecasting,* vol. 37, no. 4, pp. 1748-1764, 2021.

[44] J. Grigsby, Z. Wang, and Y. Qi, "Long-range Transformers for dynamic spatiotemporal forecasting," *arXiv preprint arXiv:2109.12218*, 2021.

[45] J. Devlin, M.-W. Chang, K. Lee, and K. Toutanova, "BERT: Pre-training of deep bidirectional transformers for language understanding," *arXiv preprint arXiv:1810.04805*, 2018.

[46] K. Jin, J. Wi, E. Lee, S. Kang, S. Kim, and Y. Kim, "TrafficBERT: Pre-trained model with large-scale data for long-range traffic flow forecasting," *Expert Systems with Applications,* vol. 186, pp. 115738, 2021.

[47] X. Ye, S. Fang, F. Sun, C. Zhang, and S. Xiang, "Meta Graph Transformer: A novel framework for spatial-temporal traffic prediction," *Neurocomputing*, 2021.

[48] S. Reza, M. C. Ferreira, J. Machado, and J. M. R. Tavares, "A multi-head attention-based transformer model for traffic flow forecasting with a comparative analysis to recurrent neural networks," *Expert Systems with Applications*, pp. 117275, 2022.

[49] F. Huang, P. Yi, J. Wang, M. Li, J. Peng, and X. Xiong, "A dynamical spatial-temporal graph neural network for traffic demand prediction," *Information Sciences,* vol. 594, pp. 286-304, 2022.

[50] Q. Liu, T. Liu, Y. Cai, X. Xiong, H. Jiang, H. Wang, and Z. Hu, "Explanatory prediction of traffic congestion propagation mode: A self-attention based approach," *Physica A: Statistical Mechanics and its Applications,* vol. 573, pp. 125940, 2021.

[51] J. Su, Z. Jin, J. Ren, J. Yang, and Y. Liu, "GDFormer: A graph diffusing attention based approach for traffic flow prediction," *Pattern Recognition Letters,* vol. 156, pp. 126-132, 2022.

[52] G. Zheng, W. K. Chai, and V. Katos, "A dynamic spatial–temporal deep learning framework for traffic speed prediction on large-scale road networks," *Expert Systems with Applications,* vol. 195, pp. 116585, 2022.

[53] K. Zhang, X. Feng, L. Wu, and Z. He, "Trajectory Prediction for Autonomous Driving Using Spatial-Temporal Graph Attention Transformer," *IEEE Transactions on Intelligent Transportation Systems*, 2022.

[54] P. Veličković, G. Cucurull, A. Casanova, A. Romero, P. Lio, and Y. Bengio, "Graph attention networks," *arXiv preprint arXiv:1710.10903*, 2017.

[55] Z. Cui, K. Henrickson, R. Ke, and Y. Wang, "Traffic graph convolutional recurrent neural network: A deep learning framework for network-scale traffic learning and forecasting," *IEEE Transactions on Intelligent Transportation Systems,* vol. 21, no. 11, pp. 4883-4894, 2019.

[56] Y. Shi, Z. Huang, S. Feng, H. Zhong, W. Wang, and Y. Sun, "Masked label prediction: Unified message passing model for semi-supervised classification," *arXiv preprint arXiv:2009.03509*, 2020.

[57] S. Kim, and H. Kim, "A new metric of absolute percentage error for intermittent demand forecasts," *International Journal of Forecasting,* vol. 32, no. 3, pp. 669-679, 2016.

[58] L. Gondara, and K. Wang, "Mida: Multiple imputation using denoising autoencoders." pp. 260-272.

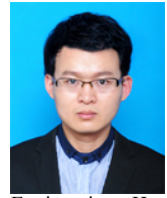

**Kunpeng Zhang** received his Ph.D. degree from the College of Mechanical and Vehicle Engineering, Hunan University, China, in 2019. From 2016 to 2017, he spent one year as a Joint Doctoral Student with the University of Michigan, Ann Arbor, Michigan, USA. He is currently a postdoctoral researcher with the Department of Automation, Tsinghua University, Beijing, and a lecturer with the College of Electrical Engineering, Henan University of Technology, Zhengzhou, China. His research interests include intelligent transportation systems and autonomous driving.

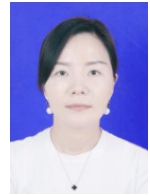

**Lan Wu** received her Ph.D. degree from the School of Automation and Information Engineering, Xi'an University of Technology, Xi'an, China, in 2009. She is currently a Professor at the College of Electrical Engineering, Henan University of Technology. Her research interests cover artificial intelligence, intelligent transportation systems, and intelligent information processing.

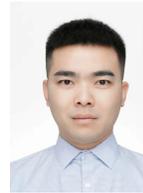

**Liang Zheng** received his Ph.D. degree from Tianjin University, Tianjin, China in 2013. From 2011 to 2012, he spent one year as a Joint Doctoral Student with the University of Wisconsin, Madison, WI, USA. Since July 2013, he has been with the School of Traffic and Transportation Engineering, Central South University, as an Associate Professor. His research interests cover traffic flow modeling and simulation, traffic prediction.

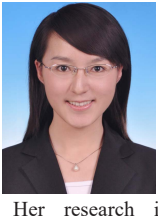

**Na Xie** received the B.S. degree from Zhejiang University, Hangzhou, China, in 2006, and the Ph.D. degree from Tsinghua University, Beijing, China, in 2011. From 2009 to 2010, she was a Visiting Scholar with the University of Cambridge, Cambridge, U.K. She is currently an Associate Professor with the School of Management Science and Engineering, Central University of Finance and Economics, Beijing, China. Her research interests include the investment and fifinancing of infrastructures, intelligent transportation investing policy, and the applications of machine learning and operational research methods, especially from economy aspects.

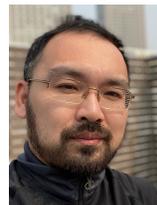

**Zhengbing He** (M'17-SM'20) received the Bachelor of Arts degree in English language and literature from Dalian University of Foreign Languages, China, in 2006, and the Ph.D. degree in Systems Engineering from Tianjin University, China, in 2011. From 2011 to 2017, he was a Postdoctoral Researcher and an Assistant Professor with the School of Traffic and Transportation, Beijing Jiaotong University, China. Presently, he is a professor of the College of Metropolitan Transportation, Beijing University of Technology, China.

His research interests include traffic flow theory, intelligent transportation systems, traffic data analytics, etc. He has published more than 120 academic papers in many mainstream transportation journals, including TR A/B/C/D, IEEE TITS/TIV, CACAIE, etc. He is an IEEE Senior Member, an Editor-in-Chief of Journal of Transportation Engineering and Information (in Chinese), an Associate Editor of IEEE TITS, an Editorial Advisory Board member of TR C. His webpage is www.GoTrafficGo.com and his email is he.zb@hotmail.com.